# Time-Aware Tranformer-Based Prediction Model for AECOPD

Weihao QU [a,1], Ling ZHENG [a], Dongyang WANG[a], Jiacun WANG[a], Haowen PAN[b]
[a] *CSSE Department, Monmouth University, West Long Branch, NJ, USA*
[b] *Changzhou Yaoyuanxing Electronic Technology Co., Ltd, China*
*ORCiD ID: Weihao Qu https://orcid.org/0000-0003-1027-6556*

**Abstract.** The rapid symptom change of Acute exacerbation of chronic obstructive pulmonary disease (AECOPD) makes it critical to have time-sensitive prediction models. However, most current machine learning models studying AECOPD use clinical and laboratory data, which will inevitably cause latency. To ensure timely detection of AECOPD and minimize latency, this paper focuses on home monitoring scenarios where only respiratory data from daily-use ventilators is available. We introduce a Time-Aware transformer-based AECOPD prediction model, which generates meaningful patient representations using the Time-Aware transformer to capture the symptoms and their temporal progression in ventilator data. Our experimental results demonstrate that our Time-Aware transformer-based approach outperforms traditional methods in multiple classification tasks, highlighting its potential to enhance AECOPD prediction accuracy.

**Keywords.** AECOPD, transformer, classification, COPD, deep learning

## 1. Introduction

Chronic Obstructive Pulmonary Disease (COPD) is a group of respiratory diseases defined by difficulty breathing, airflow limitation, and chronic cough [1]. The acute exacerbation of chronic obstructive pulmonary disease (AECOPD), refers to a sudden worsening of respiratory symptoms in excess of normal day-to-day variations [2]. AECOPD not only imposes a significant burden on patients' quality of life but also leads to increased rates of hospitalization and mortality [3]. As a result, early diagnosis and prognosis of AECOPD have become critical in clinical practice.

Machine learning research on the prediction of AECOPD relies more on symptom and clinical data. In the study of the prediction of first-time AECOPD cases [4], clinical data including electronic medical records, laboratory results, clinical visit details, and symptoms such as cough, wheezing are used as features to train machine learning models. To work on early detection of AECOPD based on the respiratory sensors data, various traditional machine learning algorithms are used, such as principal component analysis (PCA) and (SVM) classifier [5], gradient-boosted decision tree (GBDT) and the logistic regression (LR) [6]. Recent studies also attempt to use deep learning such as the feed-forward multi-layer neural network [7], Long Short Term Memory (LSTM) [8], and interpretable models [9]. Nevertheless, these aforementioned predictive models face a common challenge: time sensitivity. AECOPD are highly sensitive to time, as symptoms

[1] Corresponding Author: Weihao Qu, wqu@monmouth.edu

Published in MEDINFO 2025 - Healthcare Smart x Medicine Deep, Studies in Health Technology and Informatics 329 (2025), 1089-1093. DOI: 10.3233/SHTI251007.

can worsen quickly, often within hours. Unfortunately, clinical and laboratory data used to train these models will inevitably cause latency. If COPD patients could receive early warnings of acute exacerbations based on data from their daily-use ventilators, timely treatment could help prevent worsening conditions and reduce hospitalizations. Inspired by this concept, we aim to develop an AECOPD prediction model specifically for home monitoring, utilizing only respiratory data from daily-use ventilators.

We come across two challenges: 1) the data overload caused by the large volumes of ventilator data generated by continuous monitoring. 2) the capability to capture the relationship between symptom deterioration indicated in the respiratory data and the time. The first challenge is solved by a novel data preprocessing approach which filters out redundant data and restructures the respiratory data, details will be discussed later. As for the second challenge, we considered using feature aggregation methods to select meaningful features from the respiratory data. However, aggregation loses the connection between symptoms and their temporal progression, makes these features hard to capture the symptom deterioration. In this paper, we choose the Time-Aware transformer [10] which uses dedicated temporal embedding to explicitly model the impact of time gaps between symptom events. The Time-Aware transformer's self-attention mechanisms enable direct connections between all time steps in a sequence, making it excellent at capturing long-range dependencies in time series ventilator data.

In summary, we propose a Time-Aware Transformer-based classification model for AECOPD, specifically designed for home monitoring scenarios. We evaluated the effectiveness of our approach in home monitoring AECOPD classification by comparing to traditional feature aggregation methods, using data from 87 COPD patient ventilators.

## 2. Data and Methodology

The data comprised continuous one-month respiratory data from 87 COPD patients, collected via daily-use ventilators between 2023 and 2025 (42 patients from 2023, 10 from 2024 and 35 from 2025). Among these, 57 patients have no acute exacerbations, while 30 patients experienced AECOPD. The daily usage duration varied between 4 to 12 hours, resulting in daily respiratory data ranging from 72,000 to 220,000 rows (5 readings per second). In addition to the exact timestamps, our data includes key patient respiratory indicators such as Flow (the standard flow rate), Pressure (the air pressure), SpO2 (blood oxygen saturation levels), and ResRate (the respiratory rate). Furthermore, our data also contains ventilation effectiveness attributes, including TidalVolume (air exhaled per cycle), MinuteVent (total air ventilated per minute), and Leak (air leakage in the system). We also removed missing value and outliers to improve the data quality for further analysis.

### *2.1. Data Preprocessing*

High-frequency data contains significant redundancy. To address the redundancy, we propose a preprocessing approach, which retains only the “jump points” within the attributes. A jump point, for a given attribute, is defined as a record where the value changes significantly compared to the subsequent record. For example, the Flow value drops from 38.40 to 9.40 with a substantial change (more than 50%) is considered as a jump point. This approach allows us to focus on significant changes, which are more important for our analysis. After preprocessing, the data is reorganized into a compact format. For each patient, we store the patient ID, label, timestamp, event type, and event value. The event type corresponds to any of the respiratory attributes, ranging from flow to leak, while the event value records the respective attribute value. This preprocessing approach reduces the data size to 1/100th of the original, compressing each patient's 30-day respiratory data from approximately 3,000,000 rows with 8 attributes to around 30,000 rows with 5 attributes.

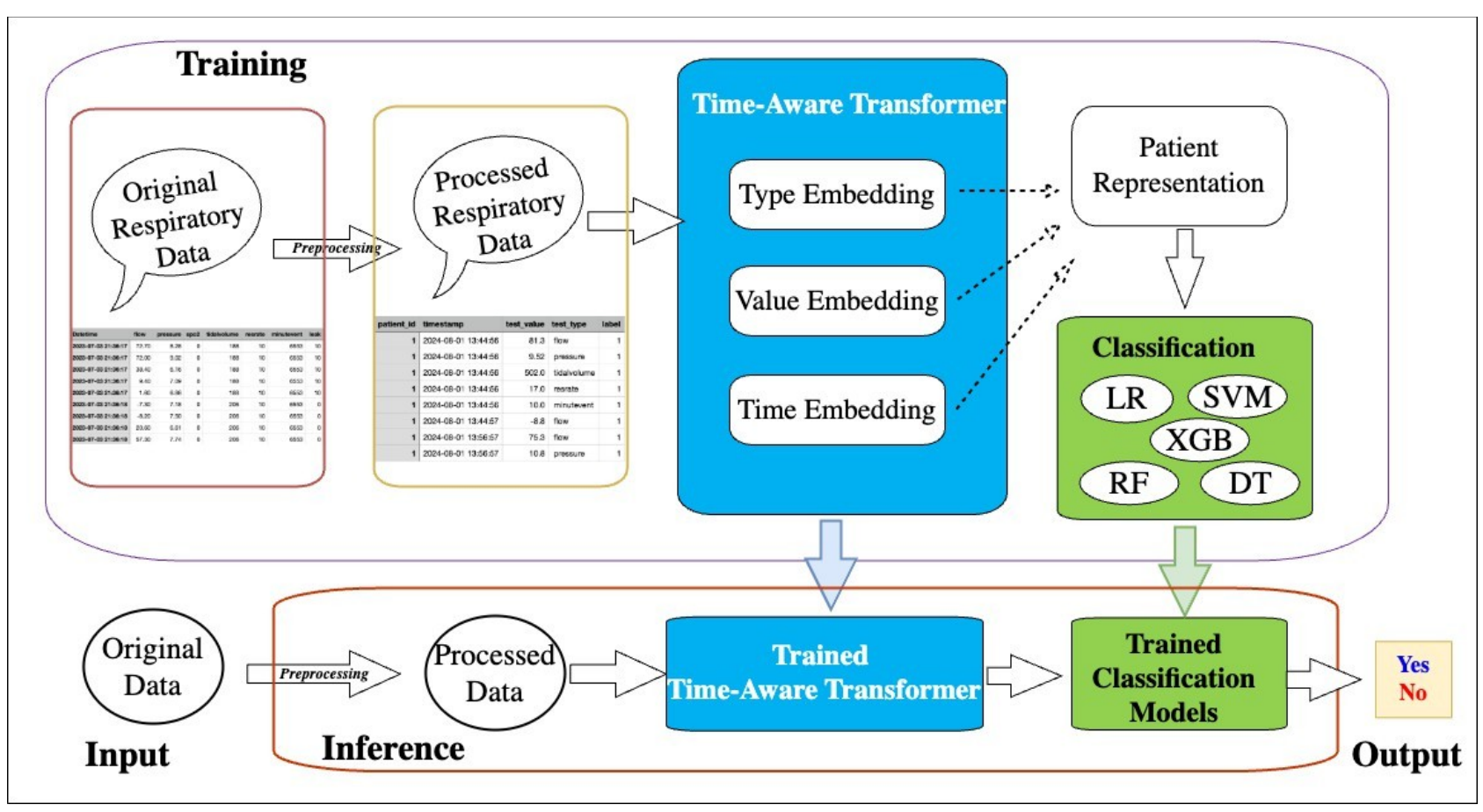

**Figure 1.** The architecture of the proposed framework

## *2.2. Methodology*

During the training stage, the preprocessed respiratory data from COPD patients is used to train a Time-Aware Transformer. The transformer generates a fixed size embedding for each patient. These embeddings are then utilized as input to train various classification models, including Logistic Regression (LR), Support Vector Machine (SVM), Random Forest (RF), Decision Tree (DT), and XGBoost (XGB). We hypothesize that the patient representation learned by the transformer from the time-series respiratory data encapsulates essential information that helps distinguish between patients with AECOPD and those without, thereby enhancing the performance of traditional classification models. In the inference stage, to predict whether a new patient is at risk of AECOPD, their preprocessed ventilator data is first converted into a patient representation by the trained transformer. The trained classification models then analyze this representation to classify the patient as at risk or not at risk of AECOPD.

## *2.3. Time-Aware Transformer*

We employ a Time-Aware Transformer to model the time-series respiratory data of COPD patients. As depicted in Figure 1, the event types and event values from the processed respiratory data are embedded into dense vector representations through the embedding layers of the transformer, resulting in type embeddings and value embeddings, respectively. To capture temporal dynamics, a sequence of time differences is computed by calculating the time gaps between consecutive events. This time difference sequence is similarly embedded into a dense vector representation, referred to as time embeddings. The transformer encoder utilizes self-attention mechanisms to manage temporal and sequential dependencies effectively. A single patient representation is then generated by aggregating information across the sequence using a masked sum of the transformer outputs, enabling the model to capture the significance of elapsed time between events, enhancing its ability to analyze and predict AECOPD occurrences.

## 3. Experiments and Discussions

This section presents empirical studies conducted to evaluate the performance of our Time-Aware Transformer-based model. To demonstrate its effectiveness, we also compare its results with classification models using traditional feature aggregation methods.

In our implementation of the Time-Aware Transformer, we use 4 attention heads, 2 transformer encoder layers, and set the size of the feedforward layers within each transformer encoder layer to 128. The loss function employed is cross-entropy loss, optimized using the Adam optimizer with a learning rate of 0.001. The maximum number of epochs is set to 200, with a consistent batch size of 2. The output of the transformer: the vectorized patient representations, are used to train various classification models. The comparative baseline uses feature aggregation methods to extract meaningful features from the original respiratory data for classification models. Two types of sequences are derived from the respiratory waveform data: FF (Peak-to-Peak Interval Sequence) and FD (First- Order Difference of Peak-to-Peak Intervals). Both the FF and FD sequences are further processed to compute a range of statistical and non-linear features, including Standard Deviation (std), Root Mean Square (rms), and Variance (var).

**Table 1.** Experimental Results of Time-Aware transformed based classification.

| **Accuracy** | **LR** | **SVM** | **DT** | **RF** | **XGB** | **Accuracy** | **LR** | **SVM** | **DT** | **RF** | **XGB** |
|---|---|---|---|---|---|---|---|---|---|---|---|
| time32 | 0.68 | 0.82 | 0.82 | 0.77 | 0.82 | notime32 | 0.58 | 0.68 | 0.59 | 0.59 | 0.64 |
| time64 | 0.73 | 0.72 | 0.68 | 0.82 | 0.82 | notime64 | 0.64 | 0.67 | 0.64 | 0.59 | 0.64 |
| time128 | 0.82 | 0.82 | 0.86 | 0.91 | 0.82 | notime128 | 0.67 | 0.59 | 0.77 | 0.59 | 0.64 |
| | | | | | | ff+fd | 0.50 | 0.50 | 0.59 | 0.63 | 0.59 |

### *3.1. Experimental Results*

We divide the 87 samples into training, validation, and test sets, resulting in 50 samples for training, 15 for validation, and 22 for testing. The test set comprises 4 samples with label 0 (no AECOPD) and 6 samples with label 1 (AECOPD). All samples are the respiratory data for 30 days.

The experimental results are shown in Table 1. We compare the accuracy of five standard classification models using patient representations generated by the Time-Aware Transformer with embedding dimensions of 32, 64, and 128 (denoted as time32, time64, and time128 in Table 1). The results indicate that an embedding dimension of 128 achieves the best performance, with accuracy scores of 0.86 and 0.91 using the Decision Tree and Random Forest classifiers, respectively. Additionally, we evaluate the impact of removing the time embedding (shown as notime32, notime64 and notime128 in Table 1), revealing that the inclusion of temporal information significantly enhances performance in distinguishing AECOPD cases. Finally, the results of the comparative baseline using FF and FD sequences (ff+fd in Table 1) demonstrate the superiority of the transformer-based approach over traditional feature aggregation methods for AECOPD classification. These findings underscore the potential of the Time-Aware Transformer for efficient and accurate prediction using only respiratory data from daily-use ventilators.

## 4. Discussion and Future Work

In this study, we explore a Time-Aware Transformer-based approach for handling multivariate time-series respiratory data from ventilators to predict AECOPD in home monitoring scenarios. Our results demonstrate that the Time-Aware Transformer excels at generating meaningful patient representations, enabling accurate AECOPD prediction. In our future work, we will exploit the use of Time-Aware Transformer in AECOPD regression task.